\documentclass[sigconf, anonymous=false, review=false, nonacm]{acmart}

\AtBeginDocument{%
  \providecommand\BibTeX{{%
    \normalfont B\kern-0.5em{\scshape i\kern-0.25em b}\kern-0.8em\TeX}}}

\setcopyright{acmcopyright}
\copyrightyear{2020}
\acmYear{2020}

\acmDOI{10.1145/1122445.1122456}

\begin{document}

\title{FaiR-N: Fair and Robust Neural Networks for Structured Data}

\author{Shubham Sharma}
\authornote{Both authors contributed equally to this research.}
\email{shubham_sharma@utexas.edu}
\affiliation{%
  \institution{The University of Texas at Austin}
  \city{Austin}
  \state{TX}
  \postcode{78704}
}

\author{Alan H. Gee}
\authornotemark[1]
\email{alangee@utexas.edu}
\affiliation{%
  \institution{The University of Texas at Austin}
  \city{Austin}
  \state{TX}
  \postcode{78704}
}

\author{David Paydarfar}
\email{david.paydarfar@austin.utexas.edu}
\affiliation{%
  \institution{The University of Texas at Austin}
  \city{Austin}
  \state{TX}
  \postcode{78704}
}

\author{Joydeep Ghosh}
\email{jghosh@utexas.edu}
\affiliation{%
  \institution{The University of Texas at Austin}
  \city{Austin}
  \state{TX}
  \postcode{78704}
}


\begin{abstract}
  Fairness in machine learning is crucial when individuals are subject to automated decisions made by models in high-stake domains. Organizations that employ these models may also need to satisfy regulations that promote responsible and ethical A.I. While fairness metrics relying on comparing model error rates across subpopulations have been widely investigated for the detection and mitigation of bias, fairness in terms of the equalized ability to achieve recourse for different protected attribute groups has been relatively unexplored. We present a novel formulation for training neural networks that considers the distance of data points to the decision boundary such that the new objective: (1) reduces the average distance to the decision boundary between two groups for individuals subject to a negative outcome in each group, i.e. the network is more fair with respect to the ability to obtain recourse, and (2) increases the average distance of data points to the boundary to promote adversarial robustness. We demonstrate that training with this loss yields more fair and robust neural networks with similar accuracies to models trained without it. Moreover, we qualitatively motivate and empirically show that reducing recourse disparity across groups also improves fairness measures that rely on error rates. To the best of our knowledge, this is the first time that recourse capabilities across groups are considered to train fairer neural networks, and a relation between error rates based fairness and recourse based fairness is investigated.
\end{abstract}




\maketitle

\section{Introduction}

Individuals are being increasingly subjected to decisions made by machine learning models in sensitive domains such as finance, law, and healthcare, where the decisions made can impact an individual in paramount ways. For such situations, it becomes essential for decisions to be unbiased, as affirmed by the General Data Protection Regulation (GDPR) \citep{butterworth2018ico}.

Fairness in machine learning is still widely debated: what makes a model fair and fair to whom? Existing fairness metrics are based on error rates (the gap between true and/or false positive rates) that are associated with different subgroups within a protected attribute. The goal is to reduce the difference between these error rates within a protected attribute to create a "fair" classifier. For example, average odds difference \citep{bellamy2018ai} reduces the sum of the differences in both true positive and false positive rates between two  groups. Equalized odds, equality of opportunity, and demographic parity \citep{barocas-hardt-narayanan} are some well-explored fairness criteria. However, there is no consensus on which form of fairness truly reflects the absence of bias, and at the very least, the suitability of each metric is context dependent \citep{mehrabi2019survey,barocas-hardt-narayanan}. 

\begin{figure}[t]
  \centering 
  \includegraphics[width = 3.1in]{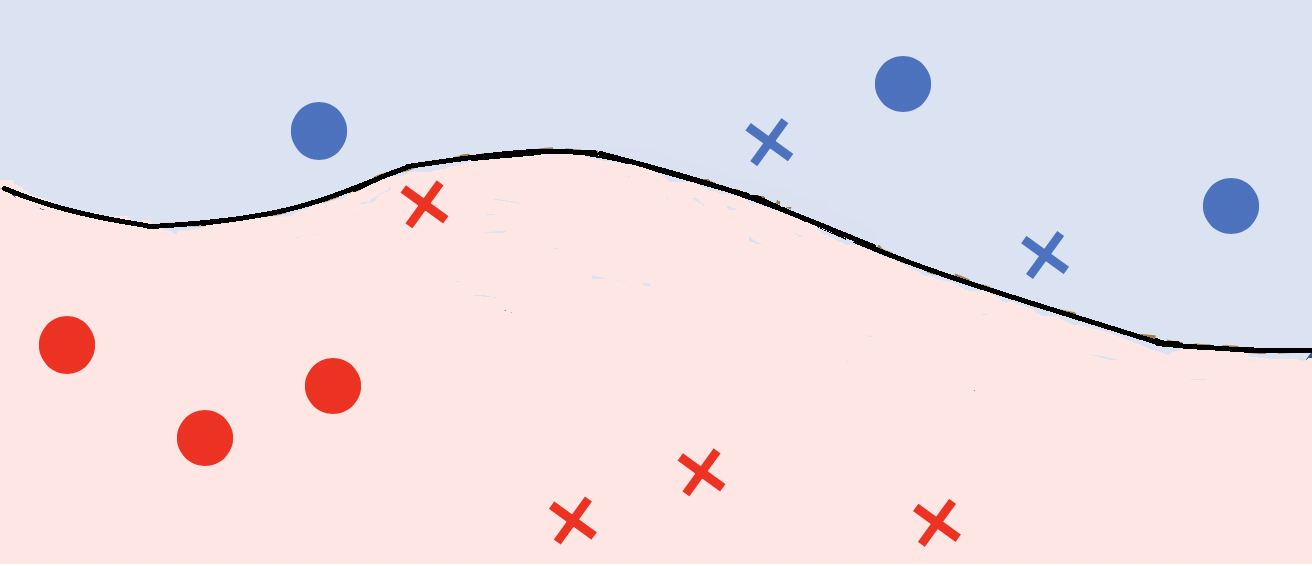} \qquad \qquad
  \includegraphics[ width=3.1in]{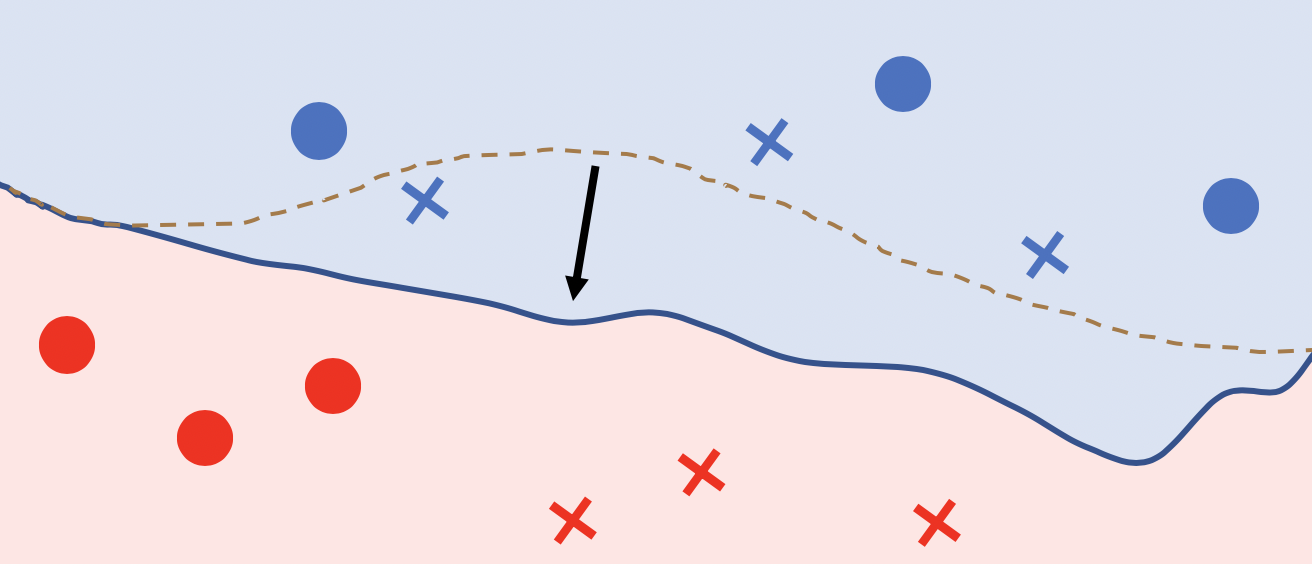}
  \caption{Depiction of the behavior of our algorithm. The red region represents negative outcome decisions, the crosses represent individuals belonging to the underprivileged group, and the circles represent individuals belonging to the privileged group. The figure on the left represents the original decision boundary. Our algorithm encourages the model to learn a decision boundary (solid line, right) that promotes fairness by reducing the difference in the difficulty of achieving recourse between the groups, while also encouraging large margin boundaries for increased robustness.}
  \label{fig:cartoon} 
\end{figure}

Notions of "distributive justice" in fairness in machine learning fail to account for the amount or nature of change that similar data profiles across different protected groups would have to make to turn a negative decision to a positive outcome. \textit{Recourse} is defined as a set of changes that need to be made to receive a positive outcome from a model. Consider a scenario where a model that promotes equalized odds denies a loan to two individuals with similar data profiles. The definition of equalized odds might show that a model is fair based on similar true and false positive rates across groups, but the ability for recourse may suggest otherwise. Without reducing recourse disparity, a female applicant may be suggested to increase her income by \$10,000 to receive a loan, while a male counterpart (with similar features) may only have to increase his income by \$5,000. Even though both individuals were granted unfavorable outcomes, the female applicant has a higher burden towards obtaining recourse - this is reflected by the distances of each data point to the model decision boundary. Hence, evaluating and ameliorating fairness by equalizing the \emph{burden of recourse} becomes important to resolve these hidden disparities.

One direction that relates to the idea of equality of recourse is the exploration of counterfactual explanations \citep{wachter2017counterfactual}, defined as the minimal change that a user can make to their input to gain a different outcome. This notion has been strongly suggested to satisfy GDPR requirements and has also been used to measure fairness through a relative measure of burden on different groups \citep{sharma2020certifai}. Since equalizing recourse closely ties in with GDPR, where individuals subjected to decisions by a model need individual level explanations, optimizing objectives based on reducing the recourse gap becomes desirable.

In this paper, we present fair and robust neural networks (FaiR-Ns) trained on structured (tabular) data that make use of a novel approximation of the distance to the decision boundary. This efficient approximation relies on the logits of the network and is then used to reduce the recourse disparity between groups. Simultaneously, the average distance of all training points (individuals) to the decision boundary is increased so that the model is more robust to perturbations, i.e. adversarial attacks. 

An example of how this loss might affect a model's decision boundary is shown in Figure \ref{fig:cartoon}. On the negative class half-space (in red) of the original model (left), the underprivileged individuals (crosses) would find it harder to obtain recourse as they are further away from the boundary on average. Moreover, some data points from the positive class are close to the boundary and can be adversarially perturbed. By utilizing the FaiR-N model (solid decision boundary on the right), the boundary is, on average, comparatively equidistant for privileged (circles) and underprivileged individuals susceptible to a negative decision (fairness). Also, on average, the boundary is distant to every data point as compared to the original model (adversarial robustness).

We demonstrate that our models achieve accuracies similar to neural networks trained with only the cross-entropy loss on three datasets from financial and healthcare domains, while reducing the burden of recourse between protected groups and promoting adversarial robustness. Any potential trade-off between promoting the two objectives (i.e. adversarial robustness and fairness with respect to minimizing recourse gap) can be explored with parameter tuning, as shown by our experiments. Bridging the recourse gap also encourages reduction of the difference between error rates across groups, and this is qualitatively motivated and empirically demonstrated. Finally, we justify the distance approximation in this paper by leveraging burden via counterfactual explanations \citep{sharma2020certifai} to verify that the networks trained with our loss reduce the recourse gap when it is measured using distances in the input space. To the best of our knowledge, this is the first time that an in-processing fairness technique for neural networks attempts to reduce recourse disparity and also investigates the relationship between disparity reduction in recourse based fairness metrics and error rates based fairness metrics.


The key contributions of this work are: 
\begin{itemize} 
    \item A novel methodology for training fair and robust neural networks using a new loss metric to induce fairness and robustness via a distance to the decision boundary formulation.
    \item A theoretical motivation and experimental validation of the relation of \textit{burden of recourse} based fairness to error rates (true and/or false positive rates) based fairness.
    \item An investigation of the trade-off between achieving recourse-based fairness and adversarial robustness during in-process model development. 
\end{itemize}

\section{Related Work}

Fairness in machine learning has been a growing field of research \citep{hacker2018teaching} that aims to mitigate the bias in existing models through techniques that can be broadly categorized into pre-processing, in-processing and post-processing methods \citep{bellamy2018ai}. Pre-processing methods manipulate the original dataset before training a model, in-processing methods modify the network itself, and post-processing techniques modify decisions made by a fixed network. However, both pre-processing and post-processing techniques face criticism for different reasons: manipulating the data leads to training a model that may not be realistic since the original distribution has been perturbed, while post-processing on a model might impose restrictions on the extent to which a model can be made fair and requires additional computation apart from training the model. In light of these concerns, in-processing techniques have been gaining traction \citep{zhang2018mitigating,mehrabi2019survey}. Surveys on pre-processing and post-processing techniques can be found in \citet{bellamy2018ai}. We focus in this section on in-processing techniques.

\citet{zhang2018mitigating} learns a classifier to maximize prediction accuracy and simultaneously reduce an adversary's ability to determine the protected attribute from the predictions. \citet{rezaei2019fair} provide an adversarial way for fair classification with logistic regression models. \citet{celis2019classification} create a meta-fair classifier with provable guarantees using simple classifiers. \citet{kamiran2010discrimination} create discrimination aware decision trees. Prejudice remover \citep{kamishima2012fairness} adds a discrimination-aware regularization term to the learning objective for a fair classifier. \citet{zemel2013learning} provide a mix of a pre-processing and in-processing techniques by learning fairer representations. \citet{zafar2017fairness} introduce disparate mistreatment that is based on misclassification rates and incorporate that into the loss functions to reduce bias. Surveys on a few other in-processing techniques that are not related to our work can be found in \citet{mehrabi2019survey}. However, none of these methods focus on reducing the recourse gap as a means to make a classifier fair, and adapting previous methods to do so is fairly non-trivial because none of them rely on the distance of a point to the boundary. Moreover, many of these methods compromise on accuracy to achieve fairness.

Adversarial robustness has been studied extensively for models trained on image and text data. Surveys on algorithms to make adversarially robust deep networks for such data can be found in \citet{ren2020adversarial}. Although \citet{ballet2019imperceptible} discuss how image adversarial attack generation and attack prevention might not generalize to tabular data, a discussion on creating models robust to adversarial attacks in tabular data is unexplored. 

Providing counterfactual explanations \citep{wachter2017counterfactual} has been strongly suggested in terms of satisfying GDPR legality requirements and has been used to measure fairness through a relative measure of burden on different groups \citep{sharma2020certifai} and provide a means of recourse \citep{spangher2018actionable}. These methods are evaluation tools and do not suggest mitigation techniques.

To estimate the distance to the boundary, \citet{elsayed2018large} use a distance approximation to the boundary to create large margin deep networks. However, their formulation is computationally expensive (a relation between this formulation and our proposed distance is established in the theory section). \citet{karimi2019characterizing} and \citet{yousefzadeh2019investigating} characterize the decision boundary by representative points on the boundary, but these methods are also computationally expensive.

\section{Theory}

We formalize the definition of fairness in the context of achieving recourse and robustness of classifiers, and then describe the distance to the boundary approximation used to apply these ideas. We also examine the relation of reducing recourse disparity to the reduction of the difference in error rates across groups. For simplicity, the theory is outlined for a binary classifier and for binary protected attributes. It can readily be extended for multiple classes and non-binary protected attributes.

\textbf{Definition 1:} A classifier is considered \textit{fair with respect to the ability to obtain recourse} when the average distance \footnote{Suitably specified (e.g. using Mahalanobis distance) to reflect cost of recourse so that the same distance implies same costs.} to the boundary for individuals subjected to a negative outcome by the classifier is equal across sub-groups defined by a protected attribute.

Specifically, given a binary classifier $\textbf{M}$ with a decision boundary $\mathcal{B}$, let $\textbf{x} \in \mathcal{X}$ be an input vector into $\textbf{M}$, let $s(\textbf{x)}$ be a stratification function that partitions the input dataset into $k$ groups based on one or more protected attributes. Let $c(\textbf{x})$ be the prediction of classifier $\textbf{M}. $ Let $d(\textbf{x},\mathcal{B})$ represent the distance to the boundary for given $\textbf{x}$. In the setting described below, we use the notation $s(\textbf{x}) \in$ $\{a,b\}$ and $c(\textbf{x}) \in$ $\{0,1\}$, where 0 represents the negative outcome and 1 represents the positive outcome. Since only individuals subject to the negative outcome need recourse, $d(\textbf{x},\mathcal{B}) =0$ for $c(\textbf{x})=1$. The fairness loss is defined as:


\begin{equation} \label{eq:fair_loss}
\mathcal{L}_{fairness} = \quad \mathop{\mathbb{| \; E}}_{\textbf{x}|{\emph{s(\textbf{x})}=a}}[d(\textbf{x},\mathcal{B})] \; -  \mathop{\mathbb{E}}_{\textbf{x}|{\emph{s(\textbf{x})}=b}}[d(\textbf{x},\mathcal{B})] \;|
\end{equation} 

The burden of recourse based fairness index is defined as:
\begin{equation}
    \mathcal{I}_{fair} = e^{-\mathcal{L}_{fairness}}
\end{equation}

Hence, a completely fair classifier has a fairness index value of 1 as $\mathcal{L}_{fairness} = 0$.

\textbf{Definition 2}: The \textit{robustness index} of a classifier is defined as the average distance of all data points to the decision boundary: 
\begin{equation}  \label{eq:robust_gain}
 \mathcal{I}_{robust} = \mathop{\mathbb{E}}_{\textbf{x}}[d(\textbf{x},\mathcal{B})] 
\end{equation}

The FaiR-N classifier introduced in this paper improves fairness and robustness while maintaining accuracy by employing a loss function that is a combination of Eq. \ref{eq:fair_loss}, Eq. \ref{eq:robust_gain}, and the standard cross-entropy loss ($\mathcal{L}_{cross}$). Specifically, FaiR-N uses the following overall objective:
\begin{equation}  \label{eq:objective}
 \mathcal{L}_{overall} = \mathcal{L}_{cross} + \lambda_{F} \cdot \mathcal{L}_{fairness} + \lambda_R  \cdot 1/\mathcal{I}_{robust}
\end{equation}
where $\lambda_{F}$ and $\lambda_R$ are hyperparameters associated with the fairness and robustness terms, respectively.

\subsection{Distance to the boundary}

Both fairness and robustness regularizers depend on the distance to the boundary for every data point. Since neural networks are non-linear, finding exact distances to the boundary can be intractable. Hence, we introduce an approximation of the distance to the boundary.

Let $f_0(\textbf{x})$ and $f_1(\textbf{x})$ be the outputs of the softmax layer of a network. Then, the decision boundary of the network can be defined as:

\begin{equation}
       \mathcal{B} = \{\textbf{x}|f_0(\textbf{x}) = f_1(\textbf{x})\}
\end{equation}

It has been shown in \citet{elsayed2018large} that based on the linearization of $f_0(\textbf{x})$ and $f_1(\textbf{x})$ around $\textbf{x}$, the distance to the boundary can be approximated as:

\begin{equation}
    d({\textbf{x},\mathcal{B}}) = \frac{ | f_0(\textbf{x}) - f_1(\textbf{x}) | }{||\nabla _\textbf{x} f_0(\textbf{x}) - \nabla _\textbf{x} f_1(\textbf{x})||_2} 
\label{eq:largedist}
\end{equation}

The above distance formulation involves gradient terms within the loss function, which requires the computation of the Hessian for optimization. To circumvent this concern, we present a more efficient approximation of the distance to boundary.
Let $g_0(\textbf{x})$ and $g_1(\textbf{x})$ represent the inputs (i.e. logits) to the softmax layer for classes $0$ and $1$, producing values $f_{0}(\textbf{x})$ and $f_{1}(\textbf{x})$, respectively. The distance of $\textbf{x}$ to the boundary is simply approximated as:

\begin{equation}  \label{eq: ourdist}
    \hat{d}({\textbf{x},\mathcal{B}}) = |g_0(\textbf{x}) - g_1(\textbf{x})|
\end{equation}

The distance approximation assumes linearity of $f(\textbf{x})$ in a local neighborhood around the decision boundary. We establish a relationship between the distance in Eq. \ref{eq:largedist} and the distance introduced in this paper (Eq. \ref{eq: ourdist}) with the following theorem:

\textbf{Theorem 1}: The relation between $d(\textbf{x},\mathcal{B})$ and $\hat d(\textbf{x},\mathcal{B})$ is monotonic and can be expressed as:

\begin{equation} \label{eq:distrel}
    d(\textbf{x},\mathcal{B}) = \frac{e^{2\hat d(\textbf{x},\mathcal{B})} - 1}{2 \; e^{\hat d(\textbf{x},\mathcal{B})} \; \nabla _\textbf{x}\hat d(\textbf{x},\mathcal{B})}.
\end{equation}

\textbf{Proof:}

Since $f_1(\textbf{x})$ and $f_0(\textbf{x}$) are outputs of the softmax activation function with logit inputs $g_0(\textbf{x})$ and $g_1(\textbf{x})$, respectively, we have the relations:

\begin{equation} \label{formofosf}
    f_1(\textbf{x}) = \frac{e^{g_1({\textbf{x})}}}{e^{g_0({\textbf{x})}} + e^{g_1({\textbf{x})}}}, \quad
    f_0(\textbf{x}) = \frac{e^{g_0({\textbf{x})}}}{e^{g_0({\textbf{x})}} + e^{g_1({\textbf{x})}}} 
\end{equation}

For a binary classifier, we have that $f_1(\textbf{x})$ = $1 - f_0(\textbf{x})$. So, taking the gradient with respect to $\textbf{x}$, we obtain:

\begin{equation} \label{eq:gradientf}
\begin{aligned}
    \nabla_x & f_1(\textbf{x}) = - \nabla_x f_0(\textbf{x}) \\
    &=Z \cdot \Big[ \nabla g_1(\textbf{x})(e^{g_0(\textbf{x})} + e^{g_1(\textbf{x})}) 
    - e^{g_1(\textbf{x})}\nabla g_1(\textbf{x}) + e^{g_0(\textbf{x})}\nabla g_0(\textbf{x}) \Big]
\end{aligned}
\end{equation}

where 

\begin{equation}
Z = \frac{e^{g_1({\textbf{x}})}}{(e^{g_0({\textbf{x}})} + e^{g_1({\textbf{x}})})^2}
\end{equation}

From Eq. \ref{eq:largedist}, we have: 

\begin{equation*}
    d(\textbf{x},\mathcal{B}) = \frac{f_0(\textbf{x}) - f_1(\textbf{x})}{2\nabla f_1(\textbf{x})} 
\end{equation*}

Substituting Eq. \ref{eq:gradientf} into Eq. \ref{eq:largedist} and simplifying, we get:

\begin{equation}
\begin{aligned}
    d(\textbf{x},\mathcal{B}) & = \frac{(e^{g_0({\textbf{x})}} - e^{g_1({\textbf{x})}})(e^{g_0({\textbf{x})}} + e^{g_1({\textbf{x})}})}{2e^{(g_0({\textbf{x}})- g_1({\textbf{x}}) )}\nabla( g_0(\textbf{x}) - g_1(\textbf{x}))} \\
& =  \frac{e^{2\hat d(\textbf{x},\mathcal{B})} - 1}{2 \; e^{\hat d(\textbf{x},\mathcal{B})} \; \nabla _\textbf{x}\hat d(\textbf{x},\mathcal{B})} 
\end{aligned}
\end{equation}

\textbf{Corollary 1}: For points near the boundary, $d(\textbf{x},\mathcal{B})$ $\simeq$ $\hat d(\textbf{x},\mathcal{B})$.

\textbf{Proof}:

For points near the boundary ($f_0(\textbf{x}) = 0.5 \pm \delta$),  $g_0(\textbf{x}) - g_1(\textbf{x})$ is small. Hence, using the first order Taylor series approximation of $e^{\textbf{x}}$, we get:

\begin{equation} \label{subnow}
\begin{split}
    d(\textbf{x},\mathcal{B}) \simeq \frac{2\hat d(\textbf{x},\mathcal{B})}{2(1+\hat d(\textbf{x},\mathcal{B}))(\nabla _\textbf{x}\hat d(\textbf{x},\mathcal{B})) }
\end{split}
\end{equation}

For points near the boundary, $f_0(\textbf{x})=0.5$. Since $f_0(\textbf{x})$ is a softmax output, $\nabla f_0(\textbf{x})=0.25$. Using Eq. \ref{formofosf} and the first order approximation of $e^{x}$:

\begin{equation}
\begin{split}
    \nabla f_0(x) \simeq \frac{1}{(1+e^{g_1(\textbf{x})-g_0(\textbf{x})})^2}(-\nabla(g_0(\textbf{x})-g_1(\textbf{x}))) \\
    = \frac{1}{(2+g_1(\textbf{x})-g_0(\textbf{x}))^2}(-\nabla(g_0(\textbf{x})-g_1(\textbf{x}))) 
\end{split}
\end{equation}

As the difference between logits is small, the denominator above would be near 2 and since $\nabla(g_0(\textbf{x})-g_1(\textbf{x})) = \hat d(\textbf{x},\mathcal{B})$, we get $\nabla _\textbf{x}\hat d(\textbf{x},\mathcal{B}) \simeq 1$.

Thus, substituting this in Eq. \ref{subnow} and using the fact that $1 > \hat d(\textbf{x},\mathcal{B})$, we obtain:

\begin{equation} \label{subnow}
\begin{aligned}
    d(\textbf{x},\mathcal{B}) & \simeq \frac{\hat d(\textbf{x},\mathcal{B})}{(1)(1) } \\
   & \simeq \hat d(\textbf{x},\mathcal{B})
\end{aligned}
\end{equation}
\qed

Empirically, we show that using both distance measures results in models with similar accuracy, fairness, and robustness characteristics. However, the distance approximation proposed in Eq. \ref{eq: ourdist} results in much faster training time, which is expected since Eq. \ref{eq:largedist} involves hessian computations.

A further justification of using our distance approximation is provided in sec 4.2 where it’s effectiveness in reducing burden as computed via an accurate but computationally expensive distance to the decision boundary in the input space based on \citet{sharma2020certifai}, is studied.

\subsection{Relation to other fairness measures}

Burden, or difficulty in achieving recourse, is zero for positive predictions and equal to the distance to the boundary for negative predictions. Burden-based fairness is then the difference in the average difficulty of recourse between groups. Hence, burden can be viewed as a soft, nuanced version of demographic parity. Specifically, if burden was fixed at 1 instead of a graded value for negative predictions, we get demographic parity.

Many machine learning models make use of error rate based fairness measures to evaluate their outcomes, e.g., disparate impact, average odds difference, equality of opportunity, and demographic parity (see \citet{barocas-hardt-narayanan} for definitions). There can be no formal guarantees on the relation of fairness via reducing recourse gap and fairness through other measures (the former relies on the distance to the boundary while the latter depend only on error rates). This can be proven as follows: Consider a perfectly accurate classifier that has an equal number of males and females in both the positive and negative class. Such a classifier would have no error rates based gap and hence, fairness notions such as demographic parity, equalized odds, and equality of opportunity would deem this classifier as fair. However, females in the negative class, on average, may still be away from the boundary compared to males. Hence, the classifier is still unfair based on the recourse based definition. Alternatively, a classifier could have males and females being equidistant from the boundary on average, but the true and false positive rates of males could still be higher, thereby imparting an unfair classifier with respect to the error rates definitions of fairness. 

Hence, while our distance approximation also encourages the reduction of error rate gaps across groups, establishing an absolute relationship between both notions of fairness is impossible. However, we theoretically motivate and empirically show that using the proposed distance approximation for the fairness loss begets improvement of other fairness measures. 


The objective for the fairness loss (Eq. \ref{eq:fair_loss}) can be simplified as:
\begin{equation}
\begin{aligned}
    \mathcal{L}&_{fairness} =  \mathop{\mathbb{| \quad E}}_{\textbf{x}|{s(\textbf{x})=a}}[d(\textbf{x},\mathcal{B})] -  \mathop{\mathbb{E}}_{\textbf{x}|{s(\textbf{x})=b}}[d(\textbf{x},\mathcal{B})] \quad | \\
     &= \big\lvert \frac{1}{n_a}\sum_{\textbf{x}|{s(\textbf{x})=a}}^{n_a} |g_1(\textbf{x}) - g_0(\textbf{x})| - 
     \frac{1}{n_b}\sum_{\textbf{x}|{s\textbf({x})=b}}^{n_b} |g_1(\textbf{x}) - g_0(\textbf{x})| \big\rvert
    \end{aligned}
\label{othfair}
\end{equation}
where $n_a$ is the number of individuals of protected attribute $a$ and $n_b$ is the number of individuals of protected attribute $b$. $g_1(x)$ and $g_0(x)$ correspond to the logits for the positive and negative class, respectively. 

Let $a$ be the privileged group, i.e. the group for which obtaining recourse on average is easier. By definition,
\begin{equation}
\begin{split}
\frac{1}{n_a} \sum_{\textbf{x}|{s(\textbf{x})=a}}^{n_a} |g_1(\textbf{x}) - g_0(\textbf{x})| < \frac{1}{n_b}\sum_{\textbf{x}|{s(\textbf{x})=b}}^{n_b} |g_1(\textbf{x}) - g_0(\textbf{x})| 
\end{split}
\label{tpfpcomp}
\end{equation}

\begin{table*}[h!]
\centering
\caption{Dataset Characteristics and Model Parameters.} \label{tab:params}
\begin{tabular}{cccccc} 
\toprule
Dataset  &   Positive Label & Protected Attribute & Training Size & Test Size & Learning Rate   \\
\midrule
Adult    & $> \$50K $ & Male/Female & 32559 & 16280 & 0.001  \\
German       & Good Credit & $>$25 years/$\le$25 years	& 800 & 200 &  0.001  \\
MEPS    & $\ge 10$ visits & White/Non-white & 7910 & 3160 & 0.0005  \\
\bottomrule
\end{tabular}
\end{table*} 

Since the individuals being considered belong to the negative class, for a reasonably accurate classifier, $g_{0}(x)$ would dominate on both sides of the equation. The recourse gap reduces between both groups with one or both of the following cases: 

\underline{Case 1}: $g_{0}(x)$ would have to increase on the left hand side of Eq. \ref{tpfpcomp} (thereby reducing $g_{1}(x)$ for the same point). This would occur if the boundary moves away from individuals belonging to the privileged class or if the boundary moves such that some individuals near the boundary that were originally classified in the positive class are now predicted to be in the negative class. This potentially increases false negatives and true negatives for the privileged class.

\underline{Case 2}: $g_{1}(x)$ would have to increase on the right hand side of Eq. \ref{tpfpcomp} (thereby reducing $g_{0}(x)$ for the same point). This would occur if the boundary moves towards individuals belonging to the underprivileged class or if the boundary moves such that some individuals near the boundary that were originally classified in the negative class are now predicted to be in the positive class, thereby potentially increasing false positives and true positives for the underprivileged class. 

Hence, minimizing  $\mathcal{L}_{fairness}$ also reduced the gaps of true and false positive rates. Since other measures of fairness rely on the disparity between error rates between groups, reducing recourse disparity with this distance approximation reduces bias with respect to other fairness measures.

\begin{figure*}[h!]
  \centering 
  \includegraphics[scale=.4]{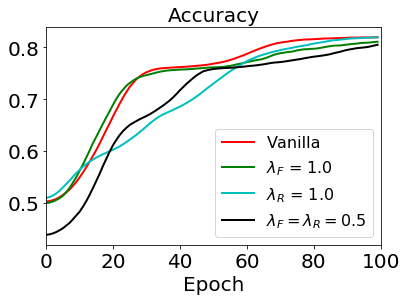}
  \includegraphics[scale=.4]{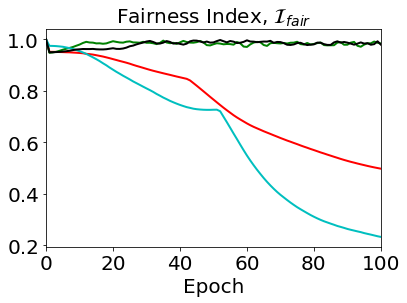}
\includegraphics[scale=.4]{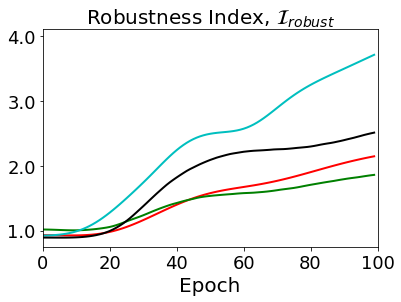} \newline

\includegraphics[scale=.38]{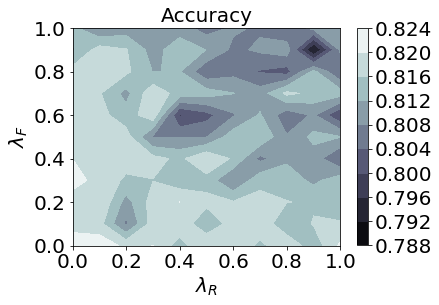}
\includegraphics[scale=.38]{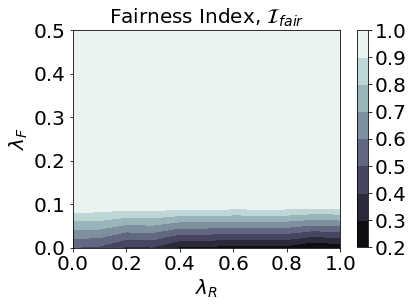}
\includegraphics[scale=.38]{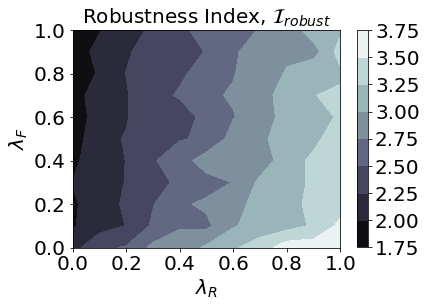}
  \caption{UCI Adult Dataset results for 100 epochs, averaged over 5 runs: (Top) Training results with gender as the protected attribute. Higher index values means better models. (Bottom) Heat maps of the test data for various regularization coefficients (with models trained on increments of 0.1 for $\lambda_{F}$ and $\lambda_{R}$)}
  \label{fig:adultvary}
\end{figure*} 

\section{Experiments and Results}

Experiments were performed with neural networks trained on three different datasets: UCI Adult \citep{kohavi1996scaling}, German Credit \citep{Dua:2019} and MEPS \citep{cohen1996medical}. The UCI Adult and German Credit datasets are datasets pertaining to the financial domain, where the goal is to predict a class of income ($>$ 50k is the positive outcome) and the classification of good or bad credit, respectively. The MEPS dataset is a medical expenditure dataset applicable to the healthcare domain, where the data is used to predict if a patient requires high utilization ($\geq$ 10 visits) for medical care or not.  We use these datasets as they are standard datasets in the fairness literature \citep{bellamy2018ai}. Gender, age, and race are the protected attributes used for the UCI Adult, German Credit, and MEPS, respectively for single attribute experiments. Both gender and race are considered for the multi-attribute experiment on the UCI Adult dataset. A two layer neural network with 30 hidden units was trained on each of the datasets for 100 epochs. Results are averaged across 5 training runs. The training-test split is given by Table \ref{tab:params}.

\subsection{Single Protected Attribute}

Gender, age, and race are the protected attributes used for the UCI Adult, German Credit, and MEPS, respectively. Neural networks with two or four hidden layers are used across datasets, as experimentation with deeper networks led to similar performance. Convolutional networks were not utilized since these structured datasets do not require such models. Models trained without the regularization terms are denoted as vanilla models. The accuracy, fairness and robustness indices, and error rate differences between groups across models on the test sets are reported.

The fairness and robustness indices significantly improve, as seen in Table \ref{Table1}. Furthermore, there is negligible loss in accuracy across all three datasets when compared to the vanilla model ($\lambda_{F} = \lambda_{R} = 0$). We also show the training evolution of the accuracy, fairness index, and robustness index with varying fairness and robustness regularization weights in Figure \ref{fig:adultvary} for the Adult dataset, which is the largest among the three datasets. The vanilla model exhibits lower fairness and robustness index values when compared to our model. This behavior suggests that individuals in the underprivileged protected group who are subjected to negative prediction decisions from the vanilla model (red) have worse recourse capabilities when compared to those same individuals in a fair classifier (cyan, green, and black) (Fig. \ref{fig:adultvary}, middle). More specifically, allowing for fairness considerations in the objective function (i.e. $\lambda_F$ = 1) results in an optimal fairness index value ($\mathcal{I}_{fair}=1$) and thus providing a fair classifier. Similarly, optimizing for robustness considerations (i.e. $\lambda_R$ = 1) leads to a model with a higher robustness index value (Fig. \ref{fig:adultvary}, right).

\begin{table*}[h!]
\centering
  \caption{Average test results across five runs. $\mathcal{I}_{fair}, \mathcal{I}_{robust}$ are the fairness and robustness indices, respectively. The bold metrics represent the best values for a dataset. Models were generated with learning rate $\gamma$ = 0.001 (* $\gamma$ = 0.0005). $^\dagger$Optimized by grid search: $\lambda_F = 0.3, \lambda_R = 0.4$.} \label{Table1}
\begin{tabular}{cccccc} 
\toprule
\multicolumn{1}{c}{} &     \multicolumn{5}{c}{Vanilla ($\lambda_{F} = \lambda_{R} = 0$)}   \\
Dataset  &   Acc. & $\mathcal{I}_{fair}$ & $\mathcal{I}_{robust}$ & $| \Delta$ TPR$|$ & $|\Delta$ FPR$|$ \\
\midrule
Adult    &  \textbf{0.821$\pm$0.002} & 0.502$\pm$0.058 & 2.16$\pm$0.052 & 0.399$\pm$0.033 & 0.105$\pm$0.007  \\
German      &  \textbf{0.767$\pm$0.007} & 0.941$\pm$0.047 & 1.43$\pm$0.061 & 0.120$\pm$0.058 & 0.215$\pm$0.078 \\
MEPS*    & \textbf{0.848$\pm$0.002} &  0.726$\pm$0.044 &       2.24$\pm$0.038 &       0.104$\pm$0.029 &       0.023$\pm$0.007  \\
\toprule
 \multicolumn{1}{c}{} &   \multicolumn{5}{c}{FaiR-N   ($\lambda_{F} = \lambda_{R} = 0.5$)}   \\
\midrule
 Adult    &  0.808$\pm$0.007 & \textbf{0.968$\pm$0.024} & \textbf{2.52$\pm$0.178} & \textbf{0.050$\pm$0.051}
 & \textbf{0.017$\pm$0.009}  \\
 German$^\dagger$      &  0.750$\pm$0.007 & \textbf{0.946$\pm$0.056} & \textbf{1.90$\pm$0.061}  & \textbf{0.053$\pm$0.037}  & \textbf{0.082$\pm$0.057}    \\
  MEPS* & 0.838$\pm$0.010 & \textbf{0.954$\pm$0.028} & \textbf{3.25$\pm$0.702} & \textbf{0.035$\pm$0.028} & \textbf{0.010$\pm$0.011} \\
\bottomrule
\end{tabular}
\end{table*}

\begin{figure*}[h!]
  \centering 
  \includegraphics[scale=.4]{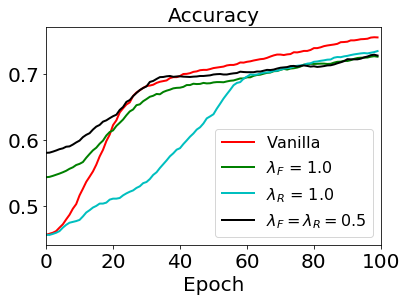}
  \includegraphics[scale=.4]{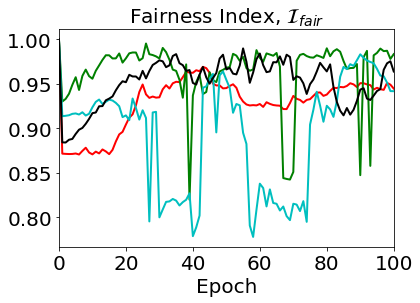} 
    \includegraphics[scale=.4]{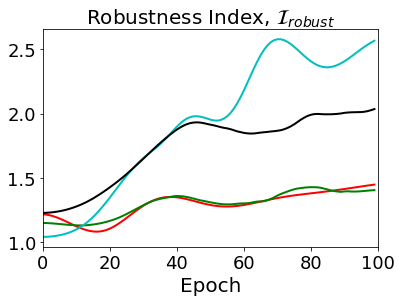}\newline
  
    \includegraphics[scale=.4]{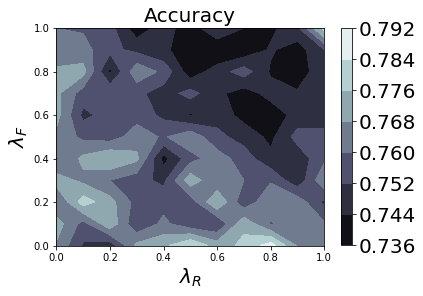}
    \includegraphics[scale=.4]{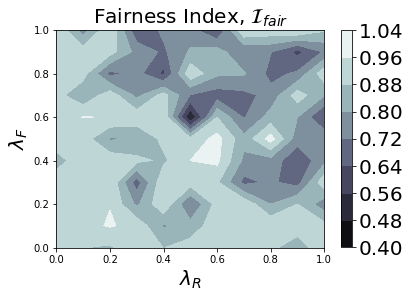}
  \includegraphics[scale=.4]{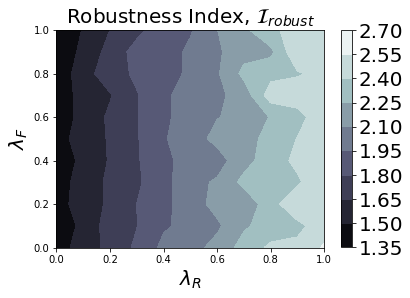}
  \caption{German Credit dataset results for 100 epochs, averaged over 5 runs: (Top) Training results with gender as the protected attribute. Higher index values means better models. (Bottom) Heat maps of the test data for various regularization coefficients (with models trained on increments of 0.1 for $\lambda_{F}$ and $\lambda_{R}$).}
  \label{fig:german_credit_plots}
\end{figure*} 

\begin{figure*}[h!]
  \centering 
  \includegraphics[scale=.4]{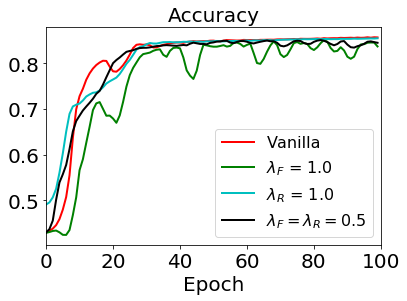}  \includegraphics[scale=.4]{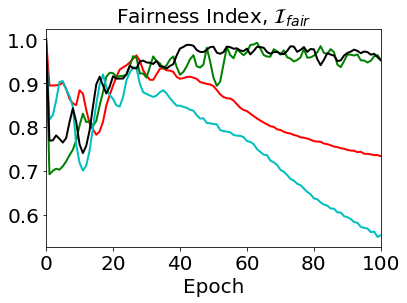} 
    \includegraphics[scale=.4]{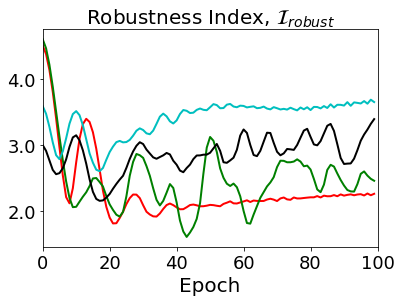} \newline

  \includegraphics[scale=.38]{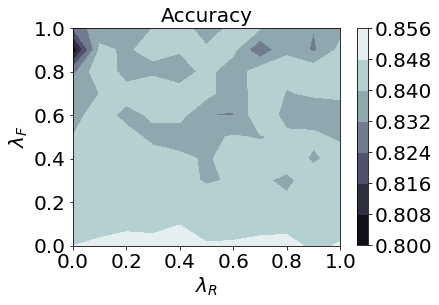}
  \includegraphics[scale=.38]{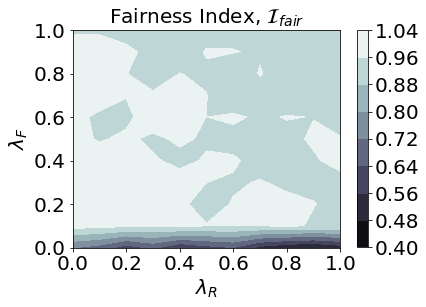}
  \includegraphics[scale=.38]{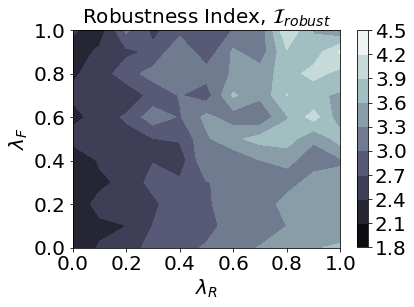}
  \caption{MEPS dataset results for 100 epochs, averaged over 5 runs: (Top) Training results with gender as the protected attribute. Higher index values means better models. (Bottom) Heat maps of the test data for various regularization coefficients (with models trained on increments of 0.1 for $\lambda_{F}$ and $\lambda_{R}$).}
  \label{fig:meps_plots}
\end{figure*}

\begin{figure*}[!h]
  \centering 
  \includegraphics[scale=.4]{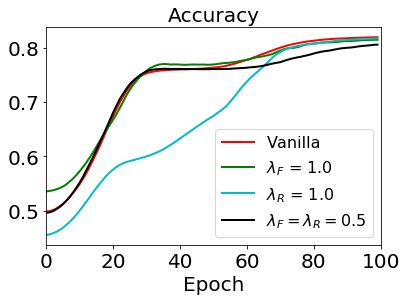} 
  \includegraphics[scale=.4]{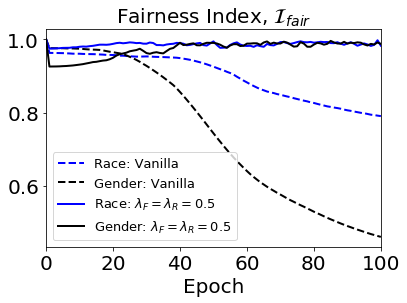}
  \includegraphics[scale=.4]{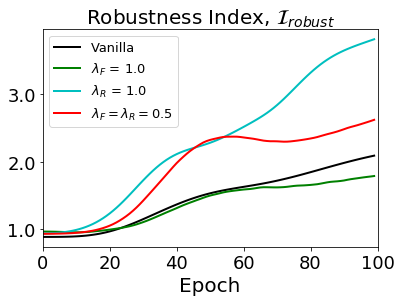}
  \caption{UCI Adult Dataset training results for models considering multiple protected attributes: gender and race. (Top Left) Accuracy: we observe minimal change in final accuracy given our fairness and robustness regularization terms; (Top Right) Fairness: we observe lower fairness loss values for each protected attribute, which represents a more fair classifier. Note that without considerations, the disparity within gender (male vs. female) and within race (white vs. black) are inherently large and unfair (i.e. vanilla model has a greater fairness loss value than FaiR-N results); (Bottom) Robustness: higher robustness index represents a more robust classifier. Increasing $\lambda_{R}$ increases the robustness of the model.}
  \label{fig:multi_adult}
\end{figure*} 

\begin{table*}[t!]
    \centering
        \caption{Comparison of models (averaged across five runs) trained on the Adult dataset from \citet{bellamy2018ai}. FaiR-N (-2 and -4 represent network sizes), LargeM \citep{elsayed2018large} are models trained using the distance in Eq. (\ref{eq:largedist}). FaiR-N is also compared to other baseline techniques$^\dagger$: prejudice remover (P.R.) \citep{kamishima2012fairness}, optimized pre-processing (O.P.P.) \citep{calmon2017optimized}, DataAug \citep{sharma2020data}, and RedApp \citep{agarwal2018reductions}. We use burden from \citet{sharma2020certifai}, calculated from distances in the input space, to compare the difference in recourse gaps ($\Delta$ Burden$_{CER}$).}
    \begin{tabular}{cccccc}
    \toprule
      Method & Accuracy & | $\Delta$TPR | & | $\Delta$FPR | & Time (s) & $\Delta$Burden$_{CER}$ \\
     \midrule
     VNN-2 & 0.82$\pm$0.09 & 0.34$\pm$0.041 & 0.13$\pm$0.01 & \textbf{318.2} $\pm$ 7.2 & 0.68$\pm$ 0.04\\
     FaiR-N-2 &  0.81$\pm$ 0.09& \textbf{0.041$\pm$ 0.067}  & \textbf{0.021 $\pm$ 0.01} & \textbf{324.4$\pm$ 8.52} & \textbf{0.06$\pm$ 0.03} \\
     FaiR-N-4 &  \textbf{0.83 $\pm$ 0.12}& 0.07 $\pm$ 0.09  & 0.05 $\pm$ 0.01 & 344.9 $\pm$ 11.92 & 0.1 $\pm$ 0.04 \\
  P.R.$^\dagger$ &  0.78$\pm$0.06 & 0.12 $\pm$ 0.03 & 0.10 $\pm$ 0.02 &371.4 $\pm$8.44 & 0.16 $\pm$ 0.05\\
     O.P.P.$^\dagger$ & 0.77$\pm$ 0.11 & 0.09 $\pm$ 0.03 & 0.04 $\pm$ 0.08 & 424.2 $\pm$ 16.10 & 0.23 $\pm$ 0.04\\
     DataAug$^\dagger$ & 0.78 $\pm$ 0.10  & 0.04 $\pm$ 0.02 & 0.03$\pm$0.01 & 520.1$\pm$13.6 & 0.22$\pm$ 0.07\\
     RedApp$^\dagger$ & 0.81 $\pm$ 0.7  & 0.05 $\pm$ 0.01 & 0.04$\pm$0.02 & 419.4$\pm$11.4 & 0.19$\pm$ 0.04\\
     LargeM-2 &  0.808 $\pm$ 0.09& 0.044 $\pm$ 0.008  & 0.020 $\pm$0.006 & 471.7 $\pm$ 19.48 & \textbf{0.03 $\pm$ 0.01} \\
     LargeM-4 &  0.829$\pm$0.12& 0.049 $\pm$ 0.004  & 0.08 $\pm$ 0.01 & 631.5 $\pm$ 19.91 & 0.10 $\pm$ 0.05 \\
     \bottomrule
    \end{tabular}
    \label{tab:comparison}
\end{table*}

There is a trade-off between the two regularization terms. Setting the  contribution from the robustness index to be too high (i.e. $\lambda_R$ to 1) may increase robustness at the cost of fairness, and vice versa. This trade-off can be seen in Figure \ref{fig:adultvary} for the Adult dataset, where for the same $\lambda_F$ regularizer, a higher value of $\lambda_R$ produces a lower fairness index. With appropriate parameter tuning (Fig. \ref{fig:adultvary}) through grid search, both the fairness and robustness indices improve while maintaining similar accuracies compared to the vanilla model. Setting the fairness regularizer to a large value results in poor accuracy, since that pushes all individuals to the positive class (as only negative individuals are considered in the fairness loss). Setting the robustness regularizer to a large value also reduces accuracy since the boundary moves away from all points.

Delving further into the reasons for this trade-off, we find that the distribution of outcomes in the Adult dataset is imbalanced as more individuals belong to the negative class. To reduce the recourse gap, the model shifts the decision boundary to bring the underprivileged individuals in the negative class closer to the boundary, creating a less adversarially robust model. When the goal is to make the model more adversarially robust and since more individuals exist in the negative class, the boundary shifts such that it moves away from the negative class decisions. Because more privileged group individuals exist in this dataset (males), the points that move from the positive to the negative class as a result of increasing the robustness index would likely belong to this group. These points would be closer to the boundary and hence, their ability to get recourse would be higher. The end result would be a fairness index that decreases since more males now have a better capability to obtain recourse.

Training experiments and hyperparameter tuning for the German Credit dataset are shown in Figures \ref{fig:german_credit_plots} and \ref{fig:meps_plots}, respectively. As we can see, the fairness and robustness index values increase while accuracies remain fairly constant. The German Credit dataset also demonstrates the potential trade-off between fairness and robustness: the fairness index training plot when both indices are included in the loss is relatively unstable, and the fairness index is low when the robustness index is high. This is because moving points towards the boundary to reduce the recourse based gap compromises on adversarial robustness. However, for both datasets, fairness and robustness is improved. Results for the MEPS dataset are shown in \ref{fig:meps_plots} where similar behavior is observed.

\subsection{Multi-Attribute Fairness}

FaiR-N has the ability to not only produce robust, fair models that account for disparities within one protected attribute, but also equalize outcomes across multiple protected attributes. Figure \ref{fig:multi_adult} depicts an example of FaiR-N that considers multiple protected attributes (race and gender) when generating a predictive model for the UCI Adult dataset, where each race and gender pair is considered as a different subgroup. 

The FaiR-N model produces a model with a decision boundary that maintains accuracy similar to the vanilla model but also minimizes disparity amongst two protected attributes considered concurrently, gender and race (Fig. \ref{fig:multi_adult}, top right). Without the fairness constraint, we see that the vanilla model produces outcomes that are unequally fair for different races, and even more-so within gender. As shown by the robustness index (Fig. \ref{fig:multi_adult}, bottom), the FaiR-N model produces a boundary more robust to adversarial attacks compared to the vanilla model. 

\subsection{Reducing the error rate gap}

We empirically validate the theoretical motivation behind the reduction of error rate gaps between groups under the proposed framework (Table \ref{Table1}). The difference in error rates for the protected attribute groups is smaller with little compromise in accuracy; the vanilla model has a significant difference in error rates while the Fair-N models have near-equalized rates.

Table \ref{tab:comparison} compares our method with several other methods that intend to improve error rate based fairness measures: (1) prejudice remover \citep{kamishima2012fairness}, which includes a fairness regularizer in logistic regression models; (2) optimized pre-processing \citep{calmon2017optimized}, which transforms data to optimize for error rate based fairness definitions; (3) DataAug, a data augmentation approach \citep{sharma2020data}; and (4) RedApp, a reductions based approach \citep{agarwal2018reductions}. We apply FaiR-N on the Adult dataset. Gender remains the protected attribute. FaiR-N takes less computation time and is more accurate. It reduces true and false positive rate gaps across groups. These values are similar or better than these other methods that were developed with the objective of reducing error rate gaps between groups.

\subsection{Evaluating the distance approximation}
To study the effects of using different distance approximations (Eq. \ref{eq:largedist} and Eq. \ref{eq: ourdist}), neural networks with two and four hidden layers are trained on the adult dataset with the loss proposed in Eq. \ref{eq:objective}, using both distance formulations. We choose to show the adult dataset for this experiment since it is the largest in size and experimentation on the other datasets yielded similar results. The results are shown in Table \ref{tab:comparison}. LargeM corresponds to using the distance approximation from Eq. \ref{eq:largedist}. Both distance formulations achieve similar results while our algorithm is faster and training time does not change significantly with network size.

The proposed distance approximation, used to compute the burden of recourse based fairness, is calculated in the logit space. Therefore, we also check if reducing this burden also reduces the burden of recourse calculated using the distance of points to the boundary in the input space. A genetic algorithm-based method suggested in  \citet{sharma2020certifai} is used to find counterfactual explanations for every input data point, and the distance of the data point to its counterfactual explanation is used as an accurate measure of the distance to the boundary. We compute the burden measure proposed in CERTIFAI \citep{sharma2020certifai} for both genders and report the difference. As shown in Table \ref{tab:comparison}, FaiR-N models have less difference in burdens for the genders when compared to other approaches even when the burden is measured using CERTIFAI. Hence, a measure of burden for recourse in the input space concurs with the burden for recourse calculated using our distance approximation. It is also interesting to note that methods built to reduce the error rate gaps also reduce the burden difference, thereby reaffirming the proposition of error rate and recourse ability based fairness measures being related.

\section{Conclusion and Future Work}
We propose a novel formulation, FaiR-N, to produce more fair and robust classification models with structured data. We introduce fairness and robustness regularizers (i.e. $\mathcal{I}_{fair}, \mathcal{I}_{robust}$) to the loss term that rely on an approximation of the distance of data points to the decision boundary during training. The resultant prediction model seeks to close the gap between the ability to get recourse for different groups and provides solutions that are more adversarially robust while also reducing the error rate differences across groups.  Furthermore, our model provides a way to adjust the  trade-off between the fairness and robustness of a model. This is achieved by varying the hyperparameter values associated with the loss - a tuning feature that does not exist in previous methodologies.

We prove that our proposed distance approximation is monotonically related to another distance formulation, but with an added benefit that our approximation reduces computational demands which enables analysis of larger, more complex datasets in faster computational time. We empirically demonstrate that the use of our approximation that calculates distances to the boundary in the logit space to measure recourse based fairness is reasonable by measuring recourse based fairness using distances in the input space using counterfactual explanations. For future work, we would investigate the possibility of generating counterfactual explanations using this notion of distance, so that individual level explanations are generated alongside model training. We would also try to adapt the distance formulation to measure recourse based on actionable changes.

\bibliographystyle{ACM-Reference-Format}
\bibliography{fairn_facct}

\appendix

\end{document}